\pdfoutput=1

\documentclass[11pt]{article}

\usepackage{ACL2023}

\usepackage{times}
\usepackage{latexsym}

\usepackage[T1]{fontenc}

\usepackage[utf8]{inputenc}

\usepackage{microtype}
\usepackage{graphicx,subfigure}
\usepackage{inconsolata}
\usepackage{multirow}
\usepackage{booktabs}
\usepackage{amsmath,amssymb}
\usepackage{bbm}
\usepackage{cleveref}
\usepackage{xspace}
\usepackage{mathtools}

\def \e{\mathbf{e}}

\def \w{\mathbf{w}}
\def \h{\mathbf{h}}

\def \y{\mathbf{y}}
\def \z{\mathbf{z}}
\def \s{\mathbf{s}}
\def \g{\mathbf{g}}

\makeatletter
\DeclareRobustCommand\onedot{\futurelet\@let@token\@onedot}
\def\@onedot{\ifx\@let@token.\else.\null\fi\xspace}
\def\eg{\emph{e.g}\onedot} 

\def\ie{\emph{i.e}\onedot}

\def\etc{\emph{etc}\onedot}

\makeatother

\usepackage{acronym}

\acrodef{nlp}[NLP]{natural language processing}
\acrodef{ie}[IE]{information extraction}
\acrodef{entr}[EntR]{entity recognition}
\acrodef{rele}[RelE]{relation extraction}
\acrodef{evente}[EventE]{event extraction}

\acrodef{Entr}[EntR]{Entity recognition}
\acrodef{Rele}[RelE]{Relation extraction}
\acrodef{Evente}[EventE]{Event extraction}

\title{Modeling Instance Interactions for Joint Information Extraction with Neural High-Order Conditional Random Field}

\author{Zixia Jia$^{1,2}$\Thanks{~This work was conducted when Zixia Jia was a research intern at BIGAI.} , Zhaohui Yan$^2$, Wenjuan Han$^3$, Zilong Zheng$^{1}$\Thanks{~Correspondence to Zilong Zheng and Kewei Tu.} , Kewei Tu$^{2\dagger}$ \\
$^1$ Beijing Institute for General Artificial Intelligence (BIGAI), Beijing, China \\
$^2$ ShanghaiTech University, Shanghai, China \\
$^3$ Beijing Jiaotong University, Beijing, China \\
\texttt{ \{jiazixia,zlzheng\}@bigai.ai} \\ 
\texttt{ \{yanzhh,tukw\}@shanghaitech.edu.cn,  wjhan@bjtu.edu.cn}}
\begin{document}
\maketitle

\begin{abstract}
Prior works on joint Information Extraction (IE) typically model instance (\eg, event triggers, entities, roles, relations) interactions by representation enhancement, type dependencies scoring, or global decoding. We find that the previous models generally consider binary type dependency scoring of a pair of instances, and leverage local search such as beam search to approximate global solutions. To better integrate cross-instance interactions, 
in this work, we introduce a joint IE framework (CRFIE) that formulates joint IE as a high-order Conditional Random Field. 
Specifically, we design binary factors and ternary factors to directly model interactions between not only a pair of instances but also triplets. Then, these factors are utilized to jointly predict labels of all instances.
To address the intractability problem of exact high-order inference, we incorporate a high-order neural decoder that is unfolded from a mean-field variational inference method, which achieves consistent learning and inference. The experimental results show that our approach achieves consistent improvements on three IE tasks compared with our baseline and prior work.
\end{abstract}

\section{Introduction}
\Ac{ie} has long been considered a fundamental challenge for various downstream natural language understanding tasks, such as knowledge graph construction and reading comprehension, \etc. The goal is to identify and extract structured information from unstructured natural language text, such that both users and machines can easily comprehend the entities, relations, and events within the text. 

\begin{figure}[t!]
    \centering
    \includegraphics[width=.9\linewidth]{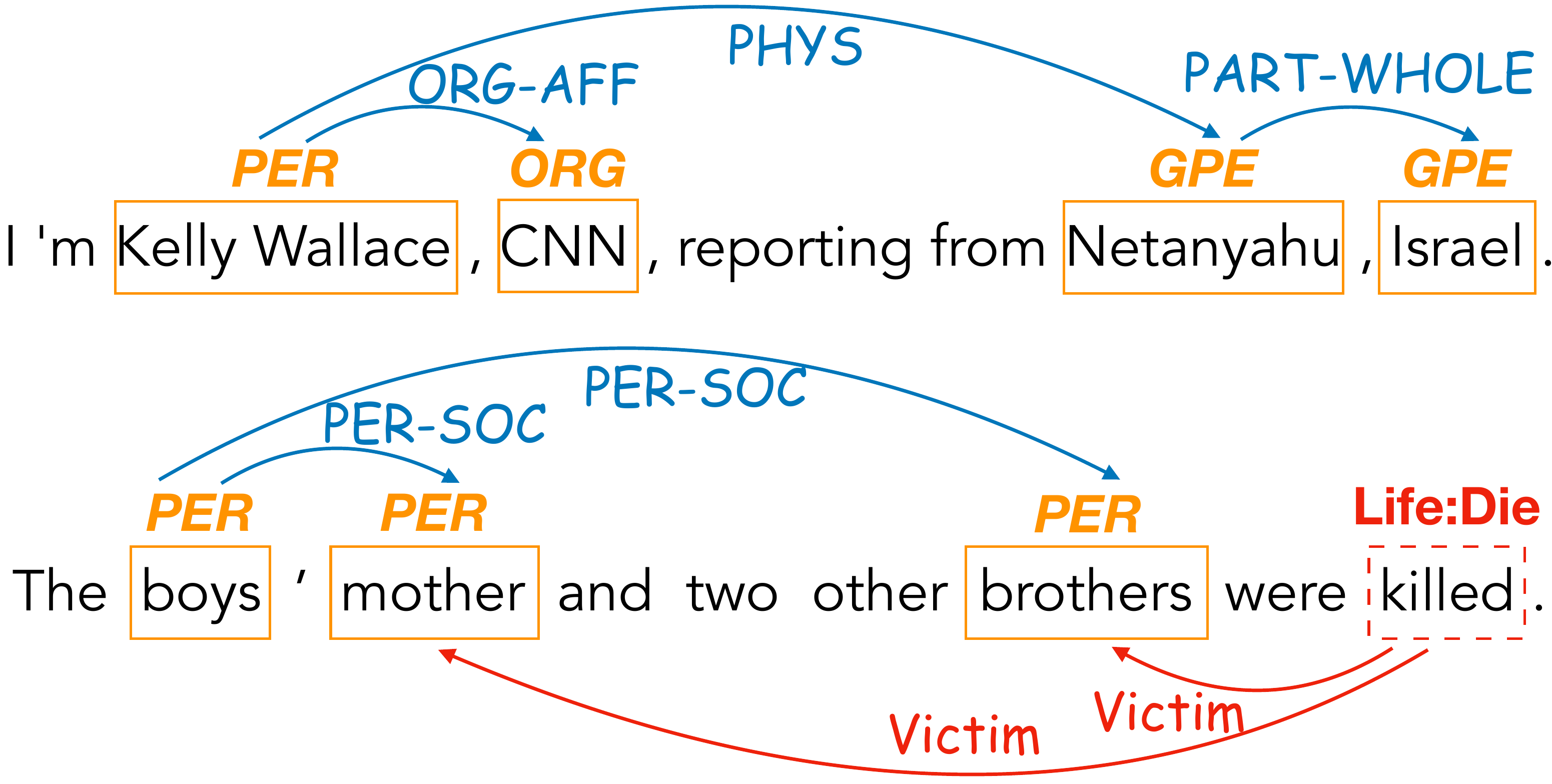}
    \caption{Example annotations of entity recognition (\eg, \smash{\includegraphics[height=9pt]{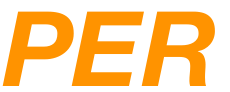}}), relation extraction (\eg, \smash{\includegraphics[height=9pt]{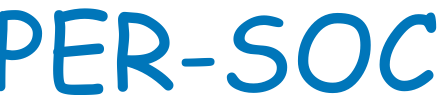}}) and event extraction tasks (\eg, \smash{\includegraphics[height=9pt]{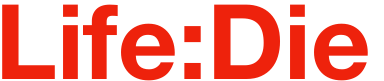}} and \smash{\includegraphics[height=9pt]{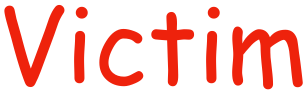}}). 
    }
    \label{fig:task}

\end{figure}

Typically, \ac{ie} consists of a series of different tasks to recognize entities, connect coreferences, extract relations, detect events, and so on. Conventional \ac{ie} schemes commonly treat different \ac{ie} tasks separately, while neglecting \textit{cross-instance} (\eg, event triggers, entities, roles, relations) or \textit{cross-task} dependencies. Such isolated learning and inference schemes lead to severely insufficient knowledge capturing and inefficient model constructions. Intuitively, predictions of different \ac{ie} instances from the same or different tasks can influence each other. For example, 
a relation between two entities would restrict the types of the entities (\eg, two entities linked by a \texttt{PART-WHOLE} relation are more likely to share entity types of the same nature, as shown in the first example of \Cref{fig:task});
types of entities can provide information that is useful to predict their relations or limit the roles they play in certain events (\eg, the knowledge of event \texttt{Life:Die} 
and entity \texttt{PER} 
can benefit the prediction of the role \texttt{Victim}, as shown in the second example of \Cref{fig:task} ). 


To effectively capture instance or task dependencies, joint \ac{ie} tries to simultaneously predict instances of different \ac{ie} tasks for an input text with a multitask learning scheme, which attracts lots of interest and demonstrates significant improvements over specific-task learning methods. Previous work of joint \ac{ie} focuses on three directions: 1) \textit{representation enrichment} by sharing the token encoder between different \ac{ie} tasks \citep{luan2018multi}, updating shared span representations according to local task-specific predictions \citep{luan2019general,wadden-etal-2019-entity}, creating dependency graphs between instances \citep{lin-etal-2020-joint, zhang2021abstract, van2021cross}, or leveraging external dependency relations such as abstract meaning representation (AMR) and syntactic structures \citep{zhang2021abstract, van2022joint};
2) \textit{type dependency scoring} 
by forming type patterns constraints \citep{lin-etal-2020-joint}, designing type dependency graphs \citep{van2021cross}, learning transition matrix of type pairs \citep{van2022joint}, or computing mutual information (MI) scores of each pair of types \citep{van2022learn};
3) \textit{global decoding} by beam search according to global features or AMR graphs \citep{lin-etal-2020-joint, zhang2021abstract}, or adopting global optimization algorithms such as simulated annealing \citep{van2022joint}. 
Our interest lies in the second and third directions and we find two main limitations of prior works. The first one is that they only score binary dependencies of instance types (\ie constraint, transition, or MI scores between a pair of types). The second one is that their decoders are based on discrete local search strategies to approximate global optima, and they often employ different approximate strategies for inference and training.

To alleviate aforementioned limitations, we propose a novel joint \ac{ie} framework, Information Extraction as high-order CRF (CRFIE), that \textit{explicitly} models label correlations between different instances from the same or different tasks, and utilizes them to calculate a joint distribution for final instance label predictions.
Specifically, we demonstrate the effectiveness of our proposed high-order framework on three widely-explored \ac{ie} tasks: \ac{entr}, \ac{rele} and \ac{evente}. We formulate the three tasks as a unified graph prediction problem, further modeled as a high-order Conditional Random field (CRF) \citep{ghamrawi2005collective}, where variables contain node variables and edge variables representing trigger/entity instances and role/relation instances respectively. The term ``high-order'' refers to factors connecting two or more correlated variables. Beyond the unary (first-order) factor, we design not only the binary (second-order) factor to model the interactions between a pair of edge variables but also the ternary (third-order) factor to model the interactions between node-edge-node variables. Since the correlated instances may come from the same or different tasks, we categorize our high-order factors into two types: \textbf{homogeneous factors (homo)} representing correlations between instances of the same task, and \textbf{heterogeneous factors (hete)} representing correlations between instances of different tasks. Taking \ac{entr} and \ac{evente} as an example, we calculate binary factor potentials of role-role pairs~(homo), and ternary factor potentials of trigger-role-entity triplets~(hete). We leverage these scores to predict the labels of all instances jointly. Since exact high-order inference is analytically intractable, we incorporate a neural decoder that is unfolded from the approximate Mean-Field Variational Inference  (MFVI) \citep{xing2012generalized} method, which achieves end-to-end training and also consistent inference and learning processes. 
Note that MFVI can be seen as a continuous relaxation for CRF inference \cite{DBLP:journals/corr/abs-2110-14759}, which can often be more effective than discrete optimization used in previous work.
Experiments on joint \ac{ie} tasks show that CRFIE achieves competitive or better performance compared with previous state-of-the-art models\footnote{The code can be found at \url{https://github.com/JZXXX/High-order-IE}.}.

\begin{figure*}[tp]
\centering

\includegraphics[width=\textwidth]{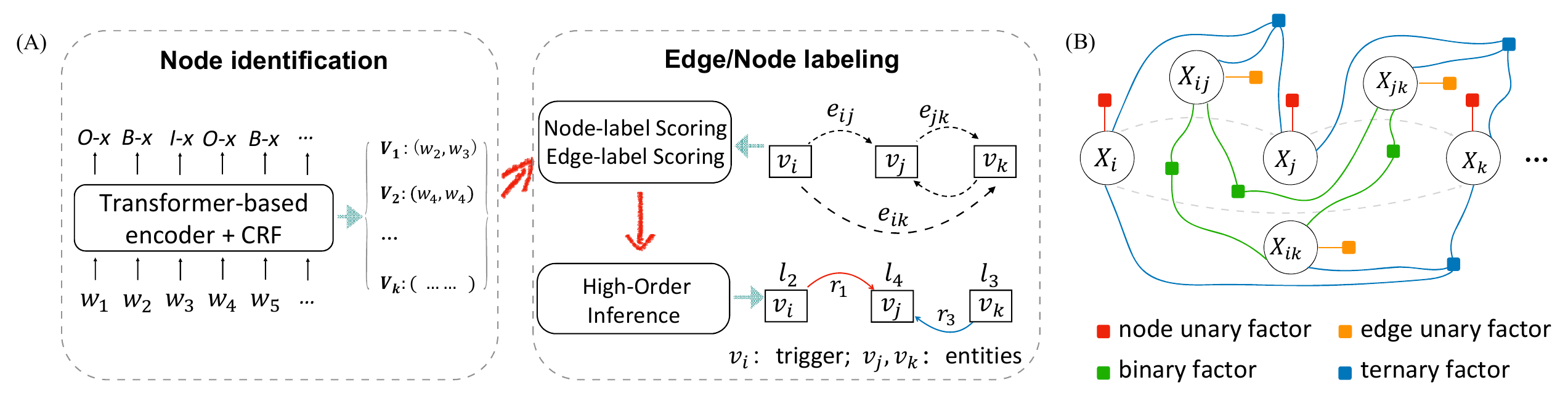}

\caption{\textbf{An overview of CRFIE.} (A) Model architecture. The identification module provides spans as nodes to the node/edge labeling module. (B) An example factor graph of our node/edge labeling module containing variables representing three nodes and three edges. $X_{i}$ indicates the label variable of the $i$-th node $v_i$ and $X_{ij}$ indicates the label variable of the edge $e_{ij}$ from the $v_i$ to $v_j$. The node labels can be event types or entity types (\ie, $X_{i}$ is the abbreviation of $X_{i}^{ntask}$ for simplicity and $ntask \in \{\text{event, entity}\}$). The edge labels can be relations or argument roles (\ie, $X_{ij}$ is the abbreviation of $X_{ij}^{etask}$ and $etask \in \{\text{relation, role}\}$). For simpler illustration, we omit edges of the opposite direction.}
\label{fig:model}

\end{figure*}

\section{Method}

\subsection{Overview of Joint IE as Graph Prediction}
We investigate three widely-explored \ac{ie} tasks.

\noindent $\rhd$ \quad EntR aims to identify some spans in a sentence as entities and label their entity types.

\noindent $\rhd$ \quad RelE aims to identify relations between some entity pairs and label their relation types.

\noindent $\rhd$ \quad EventE aims to label event types and its trigger words, identify some entities as event arguments and label argument roles.

We formulate the three \ac{ie} tasks as a graph $G = (V,E)$ prediction task, where $V$ denotes the node set 
and $E$ denotes the directed edge set. 
Each node $v = (a, b, l) \in V$ is a span for a trigger or an entity, where $a$ and $b$ index the start and end words of the span, and $l \in \mathcal{L}^{\text{event}}$ or $l \in \mathcal{L}^{\text{entity}}$ denotes the node's event type or entity type, respectively.
Each edge $e_{ij} = (i,j,r) \in E$ represents the relationship from node $v_i$ to node $v_j$,
and $r \in \mathcal{R}^{\text{role}}$ or $r \in \mathcal{R}^{\text{relation}}$ represents the edge label which is a role type when the edge is from a trigger to an entity (as an argument) or a relation type when the edge is from one entity to another.

\Cref{fig:model}(A) depicts the overall architecture of CRFIE. Because joint identification and classification need to enumerate all possible spans as nodes and high-order inference whose complexity is related to the node number becomes too computationally expensive in this situation, we follow previous work \citep{lin-etal-2020-joint,zhang2021abstract, van2021cross, van2022joint} and adopt the following pipeline: first extracting graph nodes with a node identification module, and then predicting labels of nodes and edges with a node/edge labeling module.

The \textbf{node identification module} aims to identify spans in the input sentence as graph nodes. This module is not the focus of our work, so we simply follow previous work \citep{lample2016neural,lin-etal-2020-joint,zhang2021abstract, van2021cross} to formulate node identification as a sequence labeling task with a BIO scheme. Specifically, after getting word features by averaging all sub-word embeddings extracted from a pre-trained transformer-based encoder, such as BERT~\citep{devlin2018BERT}, we use two vanilla linear-chain conditional random field~(CRF) \cite{lafferty2001conditional} as decoders to acquire trigger nodes and entity nodes separately.
We follow the conventional joint \ac{ie} settings without considering nested spans. More advanced methods such as \citet{yu-etal-2020-named, Lou2022NestedNE} can be adopted to identify graph nodes if span nesting needs to be considered. More details about the identification module can be found in Appendix \ref{sec:crf}. The identification module is fixed during subsequent training of the node/edge labeling module.

The \textbf{node/edge labeling module} is designed to predict (i) an event type for each trigger node and an entity type for each entity node and (ii) a role type for each edge between a trigger-entity pair and a relation type for each edge between an entity-entity pair. We use a special NULL label to represent non-existence of an edge. We formulate the node/edge labeling module as a high-order CRF, illustrated as a factor graph in \Cref{fig:model}(B). 
There are three kinds of factors: \textit{unary factors} that reflect the likelihood of each variable's label; \textit{binary factors} for pairs of edges sharing an endpoint, which models correlations between edge variables; and \textit{ternary factors} for an edge, its head node and its tail node, which models correlations between related node and edge variables. The joint probability over all the variables is proportional to the exponentiated sum of all the score function values of such factors. Due to the intractability of exact high-order inference, we use
MFVI to approximate it. A multitask learning scheme is adopted to train our node/edge labeling module. We describe the scoring functions, high-order inference, and learning method in the following subsections in detail.

\subsection{Unary Scoring}
\label{sec:unary}
We first obtain each node's representation $\z$ by averaging the representations of all the words within a span, in which the words' representations are obtained in the same way as in the identification module, but from another pre-trained transformer-based encoder. 
Then, the unary scores of the $i$-th node labels $\s_i^{\text{u-}\textit{ntask}} \in \mathbb{R}^{|\mathcal{L}^{\textit{ntask}}|}$ can be obtained by feeding $\z_i$ into a two layers task-specific feed-forward neural network~(FNN): 
\begin{equation}
\label{eq:u-node}
 \s^{\text{u-}\textit{ntask}}_i \! = \! \text{FNN}^{\textit{ntask}}(\z_i),
\end{equation}

where $\mathcal{L}^{\textit{ntask}}$
represents a task-specific node label set, and $ntask \in \{\text{event, entity}\}$. 

The unary scores $\s^{\text{u-}\textit{etask}}_{ij}$ of an edge $e_{ij}$ from $v_i$ to $v_j$ can be computed with a decomposed biaffine function:
\begin{equation}
\label{eq:u-edge}
    \s^{\text{u-}\textit{etask}}_{ij} \! = \! (\text{FNN}^{
\textit{etask}\text{-s}}(\z_i) \circ \text{FNN}^{\textit{etask}\text{-e}}(\z_j))\mathbf{H^{\text{u-}\textit{etask}}} \nonumber
\end{equation}
where two task-specific $\text{FNN}$s are single-layer, $\mathbf{H^{\text{u-}\textit{etask}}} \!\in \!\mathbb{R}^{d_{\textit{etask}} \times |\mathcal{R^\textit{etask}}|}$ is parameters, $\mathcal{R^{\textit{etask}}}$ represents a task-specific edge label set that includes an additional NULL label, $etask \in \{\text{relation, role}\}$, and $\circ$ denotes element-wise product.

\subsection{Binary Scoring}
\label{sec:binary}
We calculate binary correlation scores of each legal edge pair that share one endpoint. As illustrated in \Cref{fig:model}(A), there are three types of binary factors \cite{wang2019second}: edge $e_{ij}$ and edge $e_{ik}$ share the head node $v_i$, producing sibling (sib); edge $e_{jk}$ and edge $e_{ik}$ share the tail node $v_k$, producing co-parent (cop); and the tail node $v_j$ of edge $e_{ij}$ is the head node of edge $e_{jk}$, producing grandparent (gp).
For each specific type of binary factor, we use different single-layer FNNs taking $\z$ as input to calculate a head representation (-s) and a tail
representation (-e) for each node. For gp factor, we additionally calculate a middle representation (-mid) for each node.
\begin{align}
    &\g^{type\text{-s}}_{i} \! = \! \text{FNN}^{type\text{-s}}(\z_i) \quad \g^{type\text{-e}}_{i} \! = \! \text{FNN}^{type\text{-e}}(\z_i)\nonumber\\
    &\g^{\text{gp-mid}}_{i} \! = \! \text{FNN}^{\text{gp-mid}}(\z_i)\quad \!type \! \in \! \{\text{sib},\text{cop},\text{gp}\} \nonumber
\end{align}
For a sib pair $\{e_{ij}, e_{ik}\}$, cop pair $\{e_{ik}, e_{jk}\}$ and gp pair $\{e_{ij}, e_{jk}\}$, suppose that the first edge has label $r_m \in \mathcal{R}^1$ and the second edge has label $r_n \in \mathcal{R}^2$, we formulate binary scores as follows:
\begin{align*}
& s_{ijkmn}^{\text{b-sib}} \! = \! \sum\nolimits_{a=1}^{d_3}  (\g^{\text{sib-s}}_{i} \! \circ \!  \g^{\text{sib-e}}_{j} \! \circ \!  \g^{\text{sib-e}}_{k} \!  \circ \! \h^\text{1}_{m} \!  \circ \!  \h^\text{2}_{n})_a  \\
     & s_{ijkmn}^{\text{b-cop}} \! = \! \sum\nolimits_{a=1}^{d_3} \! (\g^{\text{cop-s}}_{i} \! \circ \! \g^{\text{cop-s}}_{j}  \! \circ \! \g^{\text{cop-e}}_{k} \! \circ \! \h^\text{1}_{m}  \! \circ \! \h^\text{2}_{n})_a \\
     & s_{ijkmn}^{\text{b-gp}} 
        \! = \! \sum\nolimits_{a=1}^{d_3}  (\g^{\text{gp-s}}_{i} \! \circ \! \g^{\text{gp-mid}}_{j} \! \circ \! \g^{\text{gp-e}}_{k}  \!\circ \! \h^\text{1}_{m} \!  \circ \! \h^\text{2}_{n})_a \nonumber
\end{align*}
where $\h^\text{1}_{m}$ is the embedding of the first edge label $r_m$ and $\h^\text{2}_{n}$ is the embedding of the second edge label $r_n$. All $\g$ and $\h$ are $d_3$-dimensional. For symmetry, $s_{ijkmn}^{\text{b-sib}}  \equiv s_{ikjnm}^{\text{b-sib}}$ and $s_{ijkmn}^{\text{b-cop}}  \equiv s_{jiknm}^{\text{b-cop}}$.

In this paper, we consider two types of homogeneous binary factors: \textbf{\textit{homo} case (i)} sib and cop representing two argument roles ($\mathcal{R}^1 = \mathcal{R}^2 =\mathcal{R}^{\text{role}}$) and \textbf{\textit{homo} case (ii)} sib, cop and gp representing two relations ($\mathcal{R}^1 = \mathcal{R}^2 = \mathcal{R}^{\text{relation}}$). We also consider one type of heterogeneous binary factors: \textbf{\textit{hete} case (i)} cop and gp where one edge label is a relation and the other is a role for joint EventE and RelE ($\mathcal{R}^1 = \mathcal{R}^{\text{relation}}, \mathcal{R}^2 = \mathcal{R}^{\text{role}}$ or $\mathcal{R}^1 = \mathcal{R}^{\text{role}}, \mathcal{R}^2 = \mathcal{R}^{\text{relation}}$).\footnote{It is rare that a trigger word serves as an argument meanwhile, and a relation edge and a role edge scarcely share the same head node, so we do not consider gp in \textbf{\textit{homo} case (i)} and sib in \textbf{\textit{hete} case (i)}.}

\subsection{Ternary Scoring}
\label{sec:ternary}
We calculate ternary correlation scores of an edge and its two endpoints. 
Similar to binary scoring, we use two new FNNs to produce representations for each possible head node and tail node respectively: 
\begin{align}
    \g^{\text{ter-s}}_{i} = \text{FNN}^{\text{ter-s}}(\z_i) \quad\quad \g^{\text{ter-e}}_{i} = \text{FNN}^{\text{ter-e}}(\z_i)\nonumber
\end{align}
For an edge with label $r_m \! \in \! \mathcal{R}$, its head node $v_i$ having label $l_p \! \in \! \mathcal{L}^s$ and its tail node $v_j$ having label $l_q \! \in \! \mathcal{L}^e$, the ternary score is calculated as:
\begin{equation}
\label{ternary score}
    s_{ijpqm}^{\text{ter}} \! = \sum\nolimits_{a=1}^{d_4} (\g^{\text{ter-s}}_{i} \circ \g^{\text{ter-e}}_{j} \circ \e^\text{ter-s}_{p}  \circ  \e^\text{ter-e}_{q}   \circ \h^\text{ter}_{m})_a 
\end{equation}
where $\h^\text{ter}_{m}$ is the embedding of label $r_m$, $\e^\text{ter-s}_{p}$ is the embedding of label $l_p$ and $\e^\text{ter-e}_{q}$ is the embedding of label $l_q$. $\g$, $\e$ and $\h$ are all $d_4$-dimensional. We consider two types of heterogeneous ternary factors: \textbf{\textit{hete} case (ii)} the ternary correlations between an event trigger, an entity, and a role for joint EventE and EntR ($\mathcal{L}^s = \mathcal{L}^{\text{event}}$, $\mathcal{R} =\mathcal{R}^{\text{role}}$ and $\mathcal{L}^e = \mathcal{L}^{\text{entity}}$ ) and \textbf{\textit{hete} case (iii)} two entities and their relation for joint RelE and EntR ($\mathcal{L}^s=\mathcal{L}^e = \mathcal{L}^{\text{entity}}$ and $\mathcal{R} =\mathcal{R}^{\text{relation}}$).

\subsection{High-Order Inference}
\label{sec:hoinf}
In contrast to first-order inference which independently predicts the value of each variable by maximizing its unary score, in high-order inference we jointly predict the values of all the variables to maximize the sum of their unary and high-order scores.
However, the exact joint inference on our factor graph is NP-hard in general. Therefore, we use Mean-Field Variational Inference (MFVI) \citep{xing2012generalized} for approximate inference. MFVI iteratively updates an approximate posterior marginal distribution $Q(X)$ of each variable $X$ based on messages from all the factors connected to it. For simplicity, we write $Q_i(l)$ and $Q_{ij}(r)$ to denote $Q(X_i=l)$ and $Q(X_{ij}=r)$ respectively.

Messages for edge variables aggregated from binary factors are calculated as:
\begin{align}
\small
    F^{(t)}_{\text{bi}}&(X_{ij}=r_m) 
    =\sum\nolimits_{k\neq i,j}\sum\nolimits_{r_n \in \mathcal{R}^2} \nonumber \\
    & \alpha_1 s_{ijkmn}^{\text{sib}}Q^{(t)}_{ik}(r_n) + 
    \alpha_2 s_{ikjmn}^{\text{cop}}Q^{(t)}_{kj}(r_n) \nonumber \\
    & +\alpha_3 \big(s_{ijkmn}^{\text{gp}} Q^{(t)}_{jk}(r_n) + s_{kijmn}^{\text{gp}} Q^{(t)}_{ki}\!(r_n)\big) \nonumber
\end{align}
where $\alpha_1, \alpha_2, \alpha_3 \in [0,1]$ are hyper-parameters controlling the scale of messages passed by the different types of binary factors.
These hyper-parameters are not part of standard MFVI and can instead be seen as part of the scoring function.

Messages for node variables and edge variables aggregated from ternary factors are calculated as:
\begin{align}
\small
    & F^{(t)}_{\text{ter}}(X_{ij}=r_m) \nonumber \\
    &\quad\quad= \sum_{l_p\in \mathcal{L}^s} \sum_{l_q \in \mathcal{L}^e }  s_{ijpqm}^{\text{ter}}Q^{(t)}_{i}(l_p) Q^{(t)}_{j}(l_q) \nonumber \\
    & F^{(t)}_{\text{ter}}(X_{i}=l_p)\nonumber \\
    &\quad\quad= \sum_{l_q \in \mathcal{L}^e}\sum_{r_m \in \mathcal{R}} s_{ijpqm}^{\text{ter}} Q^{(t)}_{j}(l_q) Q^{(t)}_{ij}(r_m) \nonumber \\
    & F^{(t)}_{\text{ter}}(X_{j}=l_q)\nonumber \\ &\quad\quad=\sum_{l_p \in \mathcal{L}^s }\sum_{r_m \in \mathcal{R}} s_{ijpqm}^{\text{ter}} Q^{(t)}_{i}(l_p)Q^{(t)}_{ij}(r_m) \nonumber 
\end{align}
The posterior $Q(X)$ is updated based on the messages as follows:
\begin{align}
    &Q_{ij}^{(t+1)}(r_m) \propto \exp\{s_{ijm}^{\text{u-}\textit{etask}} \nonumber \\
    &\quad+\alpha_4 F^{(t)}_{\text{bi}}\!(X_{ij}=r_m) +\alpha_5 F^{(t)}_{\text{ter}}\!(X_{ij}=r_m) \} \nonumber \\
   &Q_{i}^{(t+1)}(l_p) \propto \exp\{s_{ip}^{\text{u-}\textit{ntask}}\! + \!\alpha_6 F^{(t)}_{\text{ter}}\!(X_{i}=l_p) \} \nonumber\\
    &Q_{j}^{(t+1)}(l_q) \propto  \exp\{s_{jq}^{\text{u-}\textit{ntask}}\! + \!\alpha_7 F^{(t)}_{\text{ter}}\!(X_{j}=l_q) \} \nonumber
\end{align}
where all $\alpha \in [0,1]$ are hyper-parameters controlling the scale of different types of messages, $s_{ijm}^{\text{u-}\textit{etask}}$ is the $m$-th element of the unary potential $\s_{ij}^{\text{u-}\textit{etask}}$,
$s_{ip}^{\text{u-}\textit{ntask}}$ is the $p$-th element of the unary potential $\s_{i}^{\text{u-}\textit{ntask}}$ and $s_{jq}^{\text{u-}\textit{ntask}}$ is the $q$-th element of $\s_{j}^{\text{u-}\textit{ntask}}$. 

There are two ways of iterative MFVI update. In the synchronous update, we update $Q(X)$ for all the variables at each step. In asynchronous update, we alternate between node variables and edge variables for $Q(X)$ update. We empirically find that asynchronous update is better than synchronous update when we use ternary factors in some cases. 

The initial distribution $Q^{(0)}$ is set by normalizing exponentiated unary potentials. After a fixed $T$ (which is a hyper-parameter) number of iterations, we obtain the posterior distribution $Q^{(T)}$. For each variable, we pick the label with the highest probability according to $Q^{(T)}$ as our prediction.


\subsection{Multitask Learning}
\label{sec:train}
Given a sentence $\w=(w_1,...,w_k)$, to train multiple IE tasks with our unified high-order node-relation prediction framework, we do multi-task learning with cross-entropy losses as follows:
 \begin{align}
 \label{eq:edgeloss}
    \mathcal{L}  = &-\sum_{\mathclap{i}}  \log P(\hat{X}_i^{ntask}|\w) \nonumber \\
                   & - \sum_{\mathclap{i,j}}  \log P(\hat{X}_{ij}^{etask}|\w) \nonumber
\end{align}
where $\hat{X}_i^{ntask}$ and $\hat{X}_{ij}^{etask}$ denote the ground truth labels of nodes and edges respectively for all the tasks. The conditional distributions over node labels and edge labels with first-order inference are 
\begin{align}
    &P(X_i^{ntask}|\mathbf{w}) = (\text{SoftMax}(\s_i^{\text{u-}\textit{ntask}}))_{X_i^{ntask}} \nonumber\\
    &P(X_{ij}^{etask}|\mathbf{w}) = (\text{SoftMax}(\s_{ij}^{\text{u-}\textit{etask}}))_{X_{ij}^{etask}}, \nonumber
\end{align}
and those with high-order inference are:
\begin{align}
    &P(X_i^{ntask}|\mathbf{w}) = Q_i^{(T)}(X_i^{ntask}) \nonumber\\
    &P(X_{ij}^{etask}|\mathbf{w}) = Q_{ij}^{(T)}(X_{ij}^{etask}).\nonumber
\end{align}
where $Q^{(T)}$ is computed with $T$ MFVI iterations. 

Inspired by \citet{zheng2015conditional,wang2019second}, we unfold the MFVI iteration steps as recurrent neural network layers parameterized by unary and high-order scores. 
As such, we obtain an end-to-end recurrent neural network for both inference and training. Doing this has an added benefit of consistent inference and training, unlike traditional CRF approaches that may rely on different approximation methods for inference and training (see for example \citet{van2022joint}).

\section{Experiments}
\label{sec:exp}


\paragraph{Datasets}
We evaluate our model on the ACE2005 corpus \cite{ace05} which provides entity, relation, and event annotations. 
Following \citet{lu-etal-2021-text2event,lin-etal-2020-joint,wadden-etal-2019-entity}, we conduct experiments on four English datasets: ACE05-R for EntR and RelE, ACE05-E for EntR and EventE, and ACE05-E+ and ERE-EN for all the three tasks, with the same data pre-processing and train/dev/test split.
There are 7 entity types, 6 relation types, 33 event types, and 22 argument roles defined in the ACE2005 corpus. 
ERE-EN dataset is extracted by combining the data from three datasets for English (i.e.,
LDC2015E29, LDC2015E68, and LDC2015E78)
that are created under Deep Exploration and
Filtering of Test (DEFT) program. It includes 7 entity types, 5 relation types, 38 event types, and 20 argument roles. 
Statistics of all datasets we used are shown in Tabel \ref{tab:data_stat}.

\begin{table}[ht!]
\small
\centering
\setlength\tabcolsep{2pt}
\begin{tabular}{l|lcccc}
\hline
                          & Split & \#Sents & \#Entities & \#Relations & \#Events \\ \hline
\multirow{3}{*}{ACE05-R}  & Train & 10,051      & 26,473     & 4,788       & -        \\
                          & Dev   & 2,424       & 6,362      & 1,131       & -        \\
                          & Test  & 2,050       & 5,476      & 1,151       & -        \\ \hline
\multirow{3}{*}{ACE05-E}  & Train & 17,172      & 29,006     & 4,664       & 4,202    \\
                          & Dev   & 923         & 2,451      & 560         & 450      \\
                          & Test  & 832         & 3,017      & 636         & 403      \\ \hline
\multirow{3}{*}{ACE05-E+} & Train & 19,216      & 47,525     & 7,152       & 4,419    \\
                          & Dev   & 902         & 3,422      & 728         & 468      \\
                          & Test  & 676         & 3,673      & 802         & 424      \\ \hline
\multirow{3}{*}{ACE05-CN} & Train & 6841      & 29657     & 7934       & 2926    \\
                          & Dev   & 526         & 2250      & 596         & 217      \\
                          & Test  & 547         & 2388      & 672         & 190      \\ \hline
\multirow{3}{*}{ERE-EN} & Train & 14736      & 39501     & 5054       & 6208    \\
                          & Dev   & 1209         & 3369     & 408        & 525      \\
                          & Test  & 1163         & 3295      & 466        & 551      \\ \hline
\end{tabular}
\caption{Datasets statistics}
\label{tab:data_stat}
\end{table}


\paragraph{Evaluation}
We use F1 scores to evaluate our model's performance as in most previous work \cite{lu-etal-2021-text2event,lin-etal-2020-joint,wadden-etal-2019-entity,zhang2021abstract}. 
For the EntR task, an entity (\textit{Ent}) is correct if both its type and offsets match a gold entity. For the RelE task, a relation (\textit{Rel}) is correct if both its type and the offsets of its two related entities match a gold relation. In addition, a strict relation evaluation (\textit{Rel+}) requires that the types of the two related entities are also correct.
A trigger is correctly identified (\textit{Trig-I}) if its offsets match a gold trigger. It is correctly classified (\textit{Trig-C}) if its corresponding event type also matches the reference trigger. An argument is correctly identified (\textit{Arg-I}) if its offsets match a gold argument and its corresponding event type is correct. It is correctly classified (\textit{Arg-C}) if its role type also matches the reference argument.
All experimental results of our approach shown in this paper are the average of three runs with different random seeds.

\paragraph{Implementation Details} 
For fair comparison with previous state-of-the-art systems, we use the BERT-large-cased model~\cite{devlin2018BERT} or RoBERTa model \citep{liu2019roberta} as our encoder for the ACE05-E and ACE05-E+ datasets, and ALBERT model~\cite{lan2019ALBERT} as the encoder for the ACE05-R dataset. 
We train our model with BertAdam optimizer\footnote{https://github.com/huggingface/transformers}.
When we use a single kind of factor, $\alpha$ is set to 1 for the used and set to 0 for others. When multiple kinds of factors are used, $\alpha$ of the used are tunable parameters.
Detailed hyper-parameter values are provided in Appendix~\ref{sec:hyper}.

\subsection{Main Results}
We take our framework with first-order inference (i.e., independently predicting the value of each variable by maximizing its unary score) as \textbf{CRFIE baseline}. It can be seen that our baseline performs better than previous work in some cases, which benefits from the biaffine function in calculating unary scores.
We experiment with different combinations of tasks.

\begin{table}[t!]
\small
\centering
\setlength\tabcolsep{3pt}
\resizebox{\linewidth}{!}{
\begin{tabular}{lccccc}
 & \textit{Ent} & \textit{Tri-I} & \textit{Tri-C} & \textit{Arg-I} & \textit{Arg-C} \\ \hline
$\text{DYGIE++ \citep{wadden-etal-2019-entity}}^\dagger$ & 89.7 & - & 69.7 & 53.0 & 48.8 \\
\citet{DBLP:journals/dint/ZhangJS19}$^\circ$ & 87.1 & 73.9 & 72.0 & 57.2 & 52.4 \\
$\text{OneIE \citep{lin-etal-2020-joint}}^\dagger$ & 90.2 & 78.2 & 74.7 & 59.2 & 56.8 \\
Text2Event \citep{lu-etal-2021-text2event}$^*$ & - & - & 71.9 & - & 53.8 \\ 
$\text{FourIE \citep{van2021cross}}^{\dagger}$ & 91.3 & 78.3 & 75.4 & 60.7 & 58  \\
$\text{FourIE \citep{van2021cross}}^{\ddagger}$& 91.6 & - & 74.9 & - & 58.7 \\
\hline
$\textbf{CRFIE baseline}^{\dagger}$  & 90.8 & 77.7 & 74.8 & 58.5 & 56.4 
\\
$\textbf{CRFIE \textit{homo} case (i)}^{\dagger}$  & 90.8 & 77.7 & 74.6 & 58.7 & 57.1 \\
$\textbf{CRFIE \textit{hete} case (ii)}^{\dagger}$  & 90.7 & 77.7 & 74.3 & 59.2 & 57.2 \\
$\textbf{CRFIE \textit{homo} case (i) + \textit{hete} case (ii)} ^{\dagger}$ & 90.6 & 77.7 & 74.3 & 59.6 & 57.5
\\
$\textbf{CRFIE baseline}^\ddagger$ & 91.5 & 77.2 & 73.6 & 60.8 & 58.1 
\\
$\textbf{CRFIE \textit{homo} case (i)}^\ddagger$ 
& 91.4 & 77.2 & 73.5 & 61.3 & 58.8 \\
$\textbf{CRFIE \textit{hete} case (ii)}^\ddagger$ 
& 91.7 & 77.2 & 73.7 & 61.9 & 59.4 \\
$\textbf{CRFIE \textit{homo} case (i) + \textit{hete} case (ii)} ^{\ddagger}$ & 91.5 & 77.2 & 73.8 & 61.9 & 59.1
\\
\hline
\multicolumn{6}{c}{\textsc{For reference}} \\ \hline
$\text{AMRIE \citep{zhang2021abstract}}^\ddagger$ & 92.1 & 78.1 & 75 & 60.9 & 58.6 \\
$\text{GraphIE \citep{van2022joint}}^\ddagger$ & 91.4 & - & 75.1 & - & 59.4 \\
\bottomrule
\end{tabular}
}
\caption{Average F1 on ACE05-E dataset.
$\circ$, $*$, $\dagger$, $\ddagger$ mean ELMo, T5-large, BERT-large-cased and RoBERTa-large encoder, respectively. 
The results of FourIE (RoBERTa) are from \citet{van2022joint}. 
The results of AMRIE and GraphIE are listed for reference because they use external resources (AMR graph and syntactic tree).
}
\label{tab:ace05e}
\end{table}



\paragraph{Joint EntR, EventE} We compare our approach under different settings and also with previous work that did not leverage gold triggers and entities. Table \ref{tab:ace05e} shows the experimental results. 
The cases in the table (\eg, \textbf{\textit{homo} case (i)}) are corresponding to the aforementioned settings in the subsections \ref{sec:binary} and \ref{sec:ternary}.
The F1 scores of \textit{Tri-I} of different settings are the same because they are produced by the same node identification module that is fixed to fairly compare our model in different settings. 

\begin{table}[t!]
\centering
\setlength\tabcolsep{1.8pt}
\resizebox{.8\linewidth}{!}{%
\begin{tabular}{lccc}
 & \textit{Ent} & \textit{Rel} & \textit{Rel+} \\ \hline
DYGIE++ \citep{wadden-etal-2019-entity}$^\dagger$ & 88.6 & 63.4 & - \\
OneIE \citep{lin-etal-2020-joint}$^\dagger$& 88.8 & 67.5 \\
\citet{wang-lu-2020-two}$^\Delta$ & 89.5 & 67.6 & 64.3 \\
$\text{PURE}_s$ \citep{zhong2020frustratingly}$^\Delta$ & 89.7 & 69.0 & 65.6 \\
UNIRE \citep{wang2021unire}$^\Delta$ & 90.2 & - & 66.0 \\
PFN \citep{DBLP:conf/emnlp/YanZFZW21}$^\Delta$ & 89.0 & - & 66.8 \\
FourIE \citep{van2021cross}$^\dagger$ & 88.9 & 68.9 & - \\
UIE \citep{lu2022unified}$^*$ & - & - & 66.1 \\
\hline
\textbf{CRFIE baseline}$^\Delta$ & 89.8 & 69.9 & 67.5 \\ 
\textbf{CRFIE \textit{homo} case (ii)}$^\Delta$  & 90.2 & 70.8 & 68.2 \\
\textbf{CRFIE \textit{hete} case (iii)}$^\Delta$ & 90.1 & 70.4 & 68.3 \\
\hline
\multicolumn{4}{c}{\textsc{For reference}} \\ \hline
GraphIE \citep{van2022joint}$^\ddagger$ &89.3 &68.5 & - \\
$\text{PURE}_c$ \citep{zhong2020frustratingly}$^\Delta$ & 90.9 & 69.4 & 67.0 \\
$\text{PL-Marker}_\text{re-eval}$ \citep{ye2022packed}$^{\Delta}$ & 91.3 & 72.5 & 70.5 \\
\bottomrule
\end{tabular}
}
\caption{Average F1 on ACE05-R dataset. 
Subscript of $\textit{re-eval}$ means re-evaluation~(Appendix \ref{sec:reimp}) using the standard evaluation method as other work.
$*$, $\dagger$, $\ddagger$ and $\Delta$ mean T5-large, BERT-large-cased, RoBERTa-large and ALBERT-XXLarg-v1, respectively. $\text{PURE}_s$ refers to the PURE model with single-sentence features.  
The results of $\text{PURE}_c$ and PL-Marker are listed for reference because they use cross-sentence features and are not directly comparable with other models. The reason for GraphIE listed for reference is the same as in Tabel \ref{tab:ace05e}.}
\label{tab:ace05r}
\end{table}

\begin{table}[t!]
\small
\centering
\setlength\tabcolsep{3pt}
\resizebox{\linewidth}{!}{
\begin{tabular}{lcccccc}
 ACE05-E+ & \textit{Ent} & \textit{Rel}& \textit{Tri-I} & \textit{Tri-C} & \textit{Arg-I} & \textit{Arg-C} \\ 
\hline 
OneIE \citet{lin-etal-2020-joint} & 89.6 & 58.6 & 75.6 & 72.8 & 57.3 & 54.8 \\
Text2Event \citet{lu-etal-2021-text2event}$^*$ & - & - & - & 71.8 & - & 54.4 \\
FourIE \citep{van2021cross} & 91.1 & 63.6 & 76.7 & 73.3 & 59.5 & 57.5 \\
UIE \citet{lu2022unified}$^*$ & - & - & - & 73.4 & - & 54.8 \\ 
GTEE-DYNPREF \citet{liu-etal-2022-dynamic} & -  & -  & -  & 74.3  & -  & 54.7  \\ \hline
\textbf{CRFIE baseline} & 90.8 & 65.3 & 77.4 & 74.6 & 60.0 & 58.1 \\
\textbf{CRFIE \textit{hete} case (i) } & 90.7 & 65.1 & 77.4 & 74.8 & 60.3 & 58.5 \\
\textbf{CRFIE all} & 90.9 & 65.8 & 77.4 & 75.5 & 60.8 & 58.8  \\
\hline
\multicolumn{7}{c}{\textsc{For reference}} \\ \hline
$\text{GraphIE \citep{van2022joint}}$ & 91.0 & 65.4 & - & 74.8 & - & 59.9 \\
\bottomrule 
\vspace*{1.5ex}\\

 ERE-EN & \textit{Ent} & \textit{Rel}& \textit{Tri-I} & \textit{Tri-C} & \textit{Arg-I} & \textit{Arg-C} \\ 
\hline 
OneIE \citet{lin-etal-2020-joint} & 86.3 & 52.8 & 66.0 & 57.1 & 43.7 & 42.1\\
\hline
\textbf{CRFIE baseline}$^\ddagger$ & 87.6 & 54.4 & 69.9 & 61.5 & 45.9 & 44.2 \\
\textbf{CRFIE all}$^\ddagger$ & 87.4 & 55.1 & 69.9 & 61.4 & 53.5 & 51.2\\
\hline
\multicolumn{7}{c}{\textsc{For reference}} \\ \hline
$\text{AMRIE \citep{zhang2021abstract}}$$\ddagger$ & 87.9 & 55.2 & 68 & 61.4 & 46.4 & 45.0 \\
\bottomrule
\end{tabular}
}
\caption{Average F1 on ACE05-E+ and ERE-EN datasets. 
$*$ means T5-large, $\ddagger$ means RoBERTa-large. Others without mark use BERT-large-cased.
The reason for reference is the same as in Table \ref{tab:ace05e}.
We do not compare FourIE and GraphIE on ERE-EN dataset because their splittings of train/dev/test are different from ours. The results of previous work on ERE-EN are from \citet{zhang2021abstract}.}
\label{tab:ace05eplus}
\end{table}


It can be seen that our high-order model performs better than our baseline in most cases for EventE, which directly shows the benefit of high-order factors.
Compared to previous SOTA, our model performs uncompetitive on \textit{Tri-I}, because we focus on the interactions of node/edge labeling, and we did not tune the hyper-parameters of the node identification module while just keeping them the same as \citet{lin-etal-2020-joint}. Even with an unsatisfactory identification module, the results of \textit{Arg-C} which is the most difficult sub-task in EventE show that CRFIE achieves consistent improvement. It is worth noting that CRFIE with learned dependencies can achieve comparable performance with those models  \citep{zhang2021abstract, van2022joint} leveraging external syntactic or semantic dependencies. It is surprising that when we use both binary factors (\textbf{\textit{homo} case (i)}
)and ternary factors (\textbf{\textit{hete} case (ii)}) in the RoBERTa setting, the performance slightly drops. The reason may be that messages from different types of factors may conflict with each other, such that training becomes more difficult. We also experiment in the case where gold triggers and entities are given, results are shown in Appendix \ref{sec:ace05-Egold}.

\paragraph{Joint EntR and RelE}
Table \ref{tab:ace05r} shows our experimental results on the ACE05-R dataset.
We can find that CRFIE performs better than most previous work and our baseline both on EntR and RelE, which demonstrates the advantage of high-order inference. 
Similar to joint EntR and EventE, our high-order model with the combination of all factors cannot achieve further improvement, so we do not show the result of this setting.

\paragraph{Joint EntR, EventE and RelE}
Table \ref{tab:ace05eplus} shows the experimental results on the ACE05-E+ and ERE-EN datasets. On ACE05-E+, we show the result of \textbf{\textit{hete} case (i)} because this setting is not included in the above experiments. \textbf{CRFIE all} means that we use all kinds of binary and ternary factors. We can find that CRFIE achieves consistent improvement in EventE and RelE. 
Due to the space limitation,  
more ablations and experimental results can be found in Appendix \ref{sec:ablation}. 

\subsection{Analysis}
\label{sec:ana}
\paragraph{High-Order Scoring}
We study two variants of our high-order scoring. \textit{Share} means that we reuse the label representations in unary scoring for high-order scoring instead of using new label representations. \textit{W/o node reps} means that we calculate high-order scores without taking node representations into account, such that the high-order scores are only dependent on the labels regardless of the underlying text spans that constituent the nodes and edges. Table \ref{tab:ace05rana1} shows the comparison results with ternary factors on the ACE05-R dataset. We can find that the performance of the two variants both drops. 

\begin{table}[t!]
\centering
\setlength\tabcolsep{5pt}
\resizebox{.7\linewidth}{!}{
\begin{tabular}{lccc}
 & \textit{Ent} & \textit{Rel} & \textit{Rel+} \\ \hline
Ours \textit{hete} (+ter) & 90.1 & 70.4 & 68.3 \\ 
\textit{Share} & 90.0 & 69.7 & 67.5 \\
\textit{W/o node reps}  & 90.1 & 70.0 & 67.7 \\
\bottomrule
\end{tabular}
}
\caption{Comparison of the results of different high-order scoring methods on ACE05-R dataset.}
\label{tab:ace05rana1}
\end{table}

\paragraph{Message Passing of Ternary Factors}
From the message passing process involving ternary factors in Sec. \ref{sec:hoinf}, we can see that messages passed to an edge come only from its two endpoints, but a node gets messages from all possible edges connected to it, which causes asymmetry messages from ternary factors, we try synchronous and asynchronous updating strategies as described in Sec \ref{sec:hoinf}. For asynchronous updating, we firstly update edge posteriors using node posteriors for the reason that the initial node posteriors are more accurate. Table \ref{tab:ace05e-ana} shows the comparison results of the two updating strategies on the ACE05-E dataset.  
We can find that asynchronous update has an advantage over synchronous update on \textit{Arg-C} but harms or keeps the performance on \textit{Tri-C}.

\begin{table}[t!]
\centering
\setlength\tabcolsep{3pt}
\resizebox{.9\linewidth}{!}{
\begin{tabular}{lccccc}
 & \textit{Ent} & \textit{Tri-I} & \textit{Tri-C} & \textit{Arg-I} & \textit{Arg-C} \\ \hline
\textit{Asyn} (BERT) & 90.9 & 77.7 & 74.3 & 59.2 & 57.2 \\
\textit{Syn} (BERT) & 90.7 & 77.7 & 74.8 & 59.2 & 56.9 \\
\hline
\textit{Asyn} (RoBERTa) & 91.7 & 77.2 & 73.7 & 61.9 & 59.4 \\
\textit{Syn} (RoBERTa) & 91.7 & 77.2 & 73.7 & 61.3 & 58.8 \\
\bottomrule
\end{tabular}
}
\caption{Comparison of the results of synchronous and asynchronous updating strategies when we use ternary factor on ACE05-E dataset.}
\label{tab:ace05e-ana}

\end{table}

\begin{table}[t!]
\centering
\setlength\tabcolsep{5pt}
\resizebox{.8\linewidth}{!}{
\begin{tabular}{lcccc}
 & \textit{baseline} & \textit{+sib} & \textit{+ter} & \textit{+sib+ter}\\ \hline
Train   & 119.3 & 119.2 & 118.4 & 107.6\\ 
Test  & 91.2 & 85.1 & 81.4 & 77.2 \\
\bottomrule
\end{tabular}
}
\caption{Comparisons of speed (sentences/second) among the baseline and high-order models.}
\label{tab:speed}

\end{table}


\begin{figure*}
    \small
    \centering
    \includegraphics[width=0.95\linewidth]{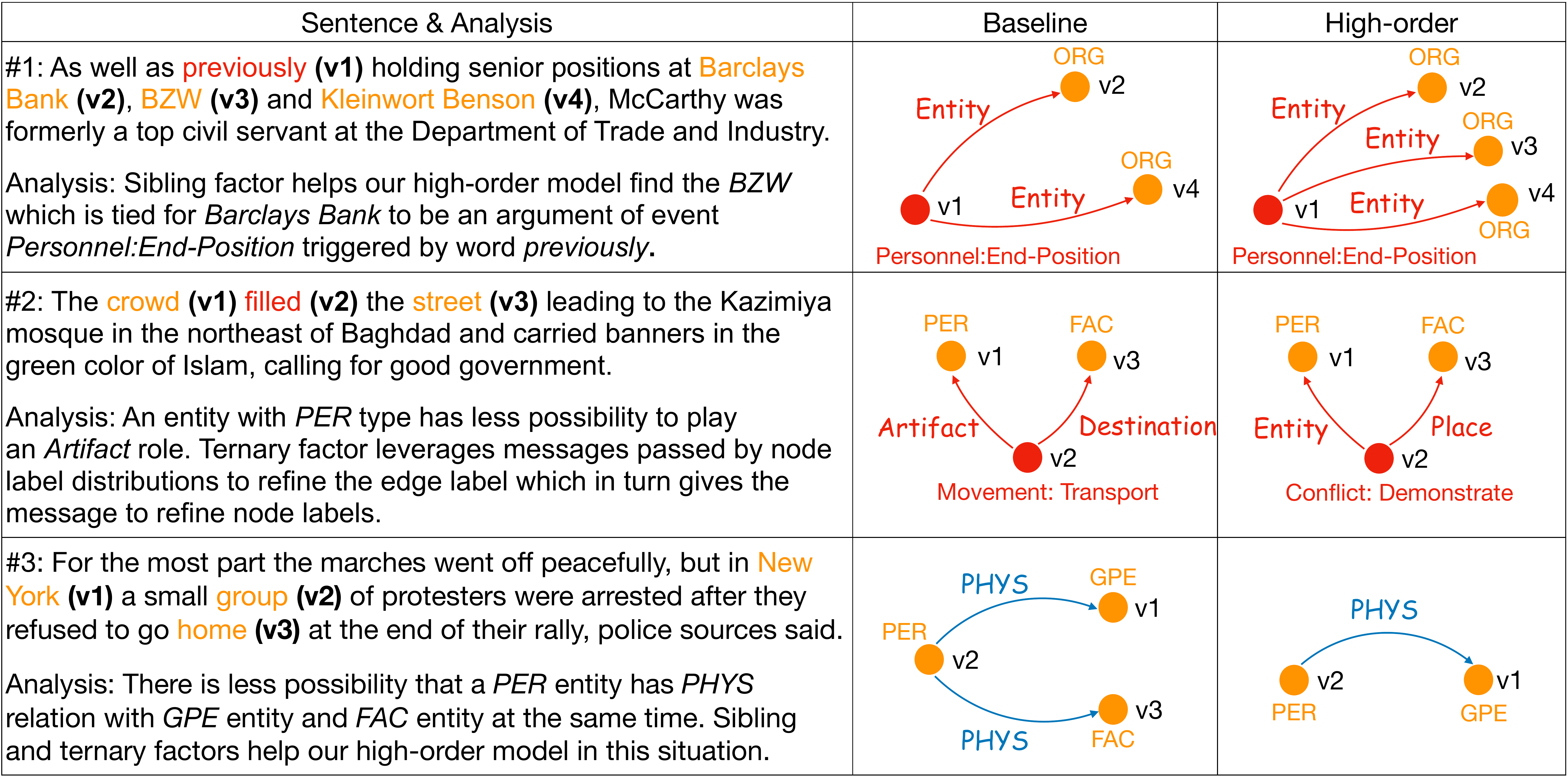}
    \caption{Examples showing how our high-order approach improves the graph prediction using different high-order factors. We only display a partial information graph for clearer illustration.}
    \label{fig:case-study}
\end{figure*}

\paragraph{Complexity and Speed of High-order Inference}
The computational complexity of our high-order inference is $O(n^3|\mathcal{R}|^2+n|\mathcal{L}|)$ when we consider binary factors and $O(n^2|\mathcal{R}||\mathcal{L}|^2)$ when we consider ternary factors, while our first-order model has a computational complexity of $O(n^2|\mathcal{R}|+n|\mathcal{L}|)$, where $n$ is the node number.  We measure the empirical training speed and inference speed on an A100 server (Table \ref{tab:speed}). We can find that our high-order models are only slightly slower than the baseline despite the difference in computational complexity, which is because we implement our models with full GPU parallelization.
\begin{figure}
    \centering
 \includegraphics[width=0.9\linewidth]{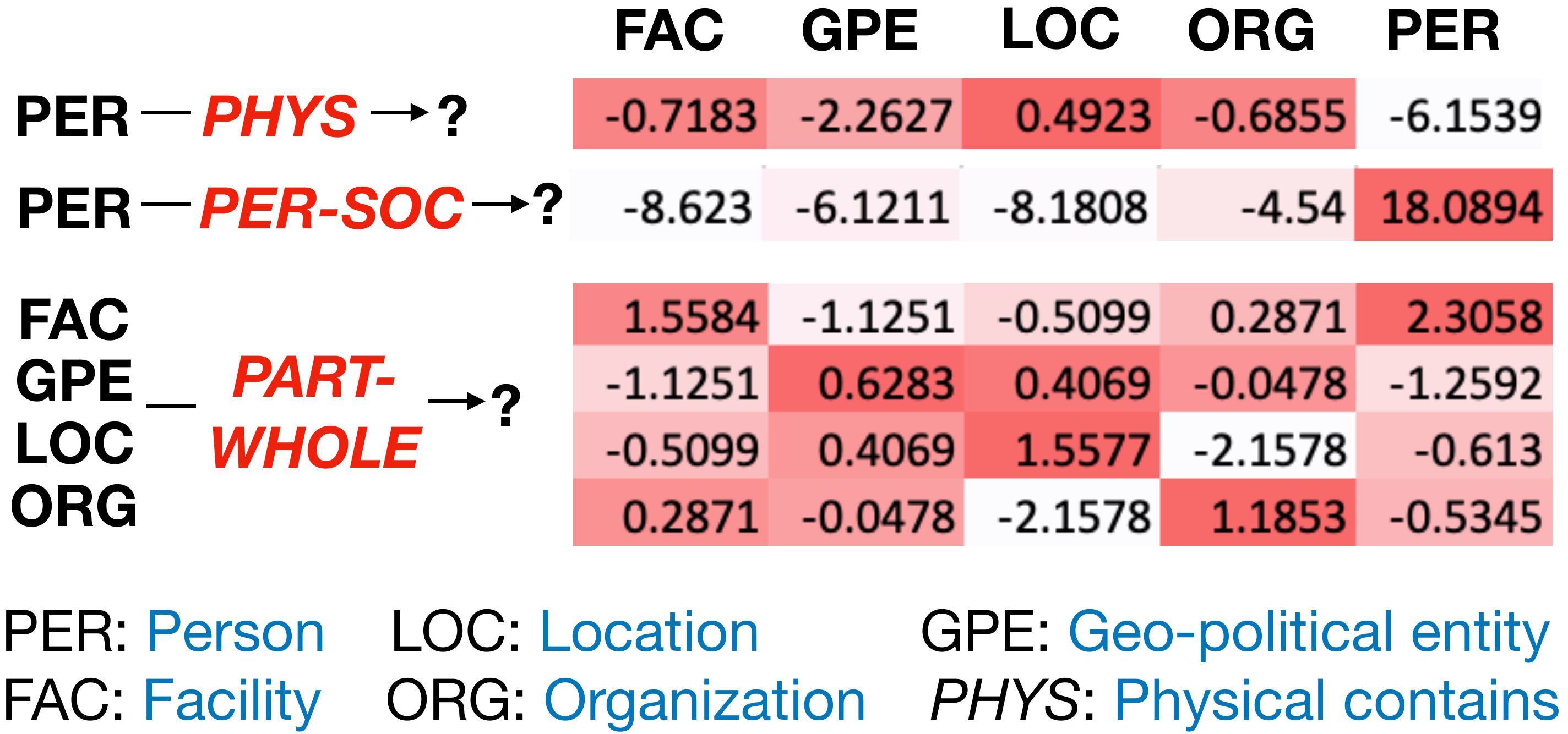}
    \caption{Ternary scores between entity-relation-entity triplets.}
    \label{fig:visul}
\end{figure}

\paragraph{Visualization of Correlation Score}
We take relation extraction as an example to visualize the ternary score calculated by Eq. \ref{ternary score} between entity-relation-entity triplets. For better understanding, we show examples of selected entity types and relation types. From Fig. \ref{fig:visul}, we can find that the correlation scores can reflect some prior knowledge. For example, `PER-SOC' relation exists between two `PER' entities, `PART-WHOLE' relation is more likely to exist between entities with the same types.

\paragraph{Error Correction Analysis and Case Study}
We provide quantitative error correction analysis in Appendix \ref{sec:error}.
Figure \ref{fig:case-study} shows examples where our high-order approach revises wrong predictions made based on the initial unary scores (i.e., the first-order baseline), along with our analyses of how high-order factors achieve the revision.

\section{Related Work}

\paragraph{Information Extraction}
Classical \ac{ie} models are typically task-specific~\citep{lample-etal-2016-neural,yu-etal-2020-named,zeng-etal-2014-relation,wang-etal-2019-extracting}. Recent efforts develop joint methods for multiple \ac{ie} tasks~\citep{DBLP:conf/emnlp/MiwaS14,DBLP:conf/acl/ZhengWBHZX17,DBLP:conf/aaai/NguyenN19,DBLP:journals/dint/ZhangJS19,wang-lu-2020-two} or general architectures for universal 
\ac{ie}~\citep{DBLP:conf/iclr/PaoliniAKMAASXS21,lu2022unified,lou2023universal}. 
Graph-based joint IE methods formulate multiple \ac{ie} tasks as a graph prediction task and aim to capture dependencies between different instances or tasks. Lots of previous works leverage encoder sharing or graph convolutional networks (GCNs) on instance dependency graphs to enrich instance representations \citep{wadden-etal-2019-entity, fu-etal-2019-graphrel, van2021cross, van2022joint, van2022learn}. This work is more relevant to some recent works that take efforts on type interactions and global inference. \citet{lin-etal-2020-joint} manually designs global features as constraints and leverages beam search to find approximated global optima. Based on the method of \citet{lin-etal-2020-joint}, \citet{van2021cross} further incorporates AMR graphs as external dependencies. The work of \citet{van2022joint} is more similar to ours in that they adopt a CRF to model type dependencies, but they learn a transition matrix that only scores binary dependencies. Besides, they employ Noise Contrastive Estimation (NCE) \citep{2013Distributed} to perform approximate training and Simulated Annealing Search to perform approximate inference. Different from their work, we model both binary and ternary dependencies and leverage MFVI to achieve consistent training and inference. 

\paragraph{High-order Methods}
Previous high-order methods most focus on instance interactions in training process to get more expressive representations, such as 
sharing representations~\citep{sun-etal-2019-joint,luan-etal-2019-general}
or using sequence-to-sequence architecture~\citep{ma-etal-2022-prompt,DBLP:conf/iclr/PaoliniAKMAASXS21,lu-etal-2021-text2event}.
There are some high-order inference methods that are related to us on different NLP tasks. 
On dependency parsing, \citet{wang2020second} considered three types of second-order parts of semantic dependencies and approximate decoding with mean-field variational inference or loopy belief propagation. \citet{jia2022span} considered interactions between two arguments of the same predicate on semantic role labeling task. However, due to the complexity, they only did high-order inference on edge existence prediction while leaving label prediction in first-order, and they did not involve heterogeneous factors. In another line of research, \citet{wang2020integrating,wang2021variational} integrate logic rules and neural network to leverage prior knowledge to help relation extraction and event extraction tasks. But they cannot achieve end-to-end training and inference.

\section{Conclusion}
In this paper, we propose a novel framework that leverages high-order interactions across different instances and different IE tasks in both training and inference processes. We formulate IE tasks as a unified graph prediction problem, further modeled as a high-order CRF.
Our framework consists of an identification module to identify spans as graph nodes and a node/edge labeling module with high-order modeling and inference to jointly label all nodes and edges. 

\section*{Limitations}
The limitation is that we separate node identification and node/edge labeling processes. Because joint node identification and label classification should enumerate all possible spans in a sentence, which is too computationally expensive. Most previous works also separate the two processes. But an obvious disadvantage of such a pipeline scheme is the error propagation problem. We take joint node identification and label classification with high-order inference as future work.

\section*{Acknowledgements}
This work is supported in part by National Key R\&D Program of China (2021ZD0150200) and the National Natural Science Foundation of China (61976139). Wenjuan Han is supported by the Talent Fund of Beijing Jiaotong University (2023XKRC006).

\bibliography{anthology,custom}
\bibliographystyle{acl_natbib}

\appendix


\section{Details on Identification Module}
\label{sec:crf}
A multi-layer perceptron (MLP) takes word representations $H=[\h_1,...,\h_n]$ as input and outputs an emission score $\mathbf{u}_i$ for each word. With a learnable transition score matrix $A$, 
a labeled sequence $\y=(y_1,...,y_n)$ can be scored as $s(\y,H)=\sum_{i=1}^{n}(\mathbf{u}_i)_{y_i}+A_{y_{i-1},y_i}$.
\paragraph{Inference} We use the Viterbi algorithm \cite{forney1973viterbi} to obtain the sequence that has the highest score: $\hat{\y} = \arg \max_{\y} s(\y,H)$.
Then we select the spans whose inter-words are labeled as B-X and I-X in the optimal output sequence as predicted node set.

\paragraph{Learning} We maximize the probability of the target sequence to learn the identification module.

\begin{small}
\begin{equation}
    P(\y^*|\w) = \frac{\exp(s(\y^*,H))}{\sum_{\y'}\exp(s(\y',H))} = \frac{1}{\mathcal{Z}}\exp (s(\y^*,H)) \nonumber
\end{equation}
\end{small}
where $\y^*$ is the target sequence and $\mathcal{Z}$ is the partition function. We can use the forward-backward algorithm \cite{dugad1996tutorial} to calculate $\mathcal{Z}$.

Of note, we did not consider nested spans in this work, which can easily be adopted to our framework using similar methods as in \citet{yu-etal-2020-named, Lou2022NestedNE} to identify graph nodes if span nesting.

\section{Hyper-parameters}
\label{sec:hyper}
For the hidden sizes of unary FNNs and most optimizer parameters, we use the default hyper-parameters following \cite{lin-etal-2020-joint}. The hidden sizes of FNNs in high-order scoring are tuned between $\{150,300\}$. The iteration step $T$ of MFVI is tuned between $\{1,2,3\}$, and it is set to 1 or 2 in different settings. We choose the hyper-parameters according to the performance of the development set after 80 epoch runs.
The main hyper-parameters are listed in Table \ref{tab:hyperparameters}.





\begin{table}[ht]
\small
\begin{tabular}{@{}lllc@{}}
\toprule
Setting  &                                    &                       & Value  \\ \midrule

\multicolumn{4}{l}{Unary   scoring}                                            \\
         & \multicolumn{2}{l}{FNN(entity)}                            & 150    \\
         & \multicolumn{2}{l}{FNN(trigger)}                           & 600    \\
         & \multicolumn{2}{l}{FNN(relation)}                          & 150    \\
         & FNN(role)                          &                       & 600    \\
\multicolumn{4}{l}{Binary   scoring}                                           \\
         & \multicolumn{2}{l}{FNN(head)}                              & 150    \\
         & FNN(tail)                          &                       & 150    \\
         & FNN(mid)                           &                       & 150    \\
\multicolumn{4}{l}{Ternary   scoring}                                          \\
         & \multicolumn{2}{l}{FNN(head)}                              & 150    \\
         & FNN(tail)                          &                       & 150    \\
\multicolumn{4}{l}{Other   setting}                                            \\
         & batch size                         &                       & 10      \\
         & \multicolumn{2}{l}{dropout rate}                           & 0.4    \\
         & \multicolumn{2}{l}{learning rate of Pretrained LM encoder} & 1e-5   \\
         & \multicolumn{2}{l}{lr decay of Pretrained LM encoder}      & 1e-5   \\
         & \multicolumn{2}{l}{learning rate of other modules}         & 1e-3   \\
         & \multicolumn{2}{l}{lr decay of other modules}              & 1e-3   \\
         & \multicolumn{2}{l}{warm-up epochs}                          & 5      \\
         & \multicolumn{2}{l}{total epochs}                          & 80      \\
         & \multicolumn{2}{l}{gradient clipping}                      & 5.0    \\ \bottomrule
\end{tabular}
\caption{Summary of hyper-parameters}
\label{tab:hyperparameters}
\end{table}

\begin{figure}[t]
\centering
    \subfigure[Our baseline]
        {                    
        \begin{minipage}[t]{0.9\columnwidth} 
        \centering                                           
        \includegraphics[width=\linewidth]{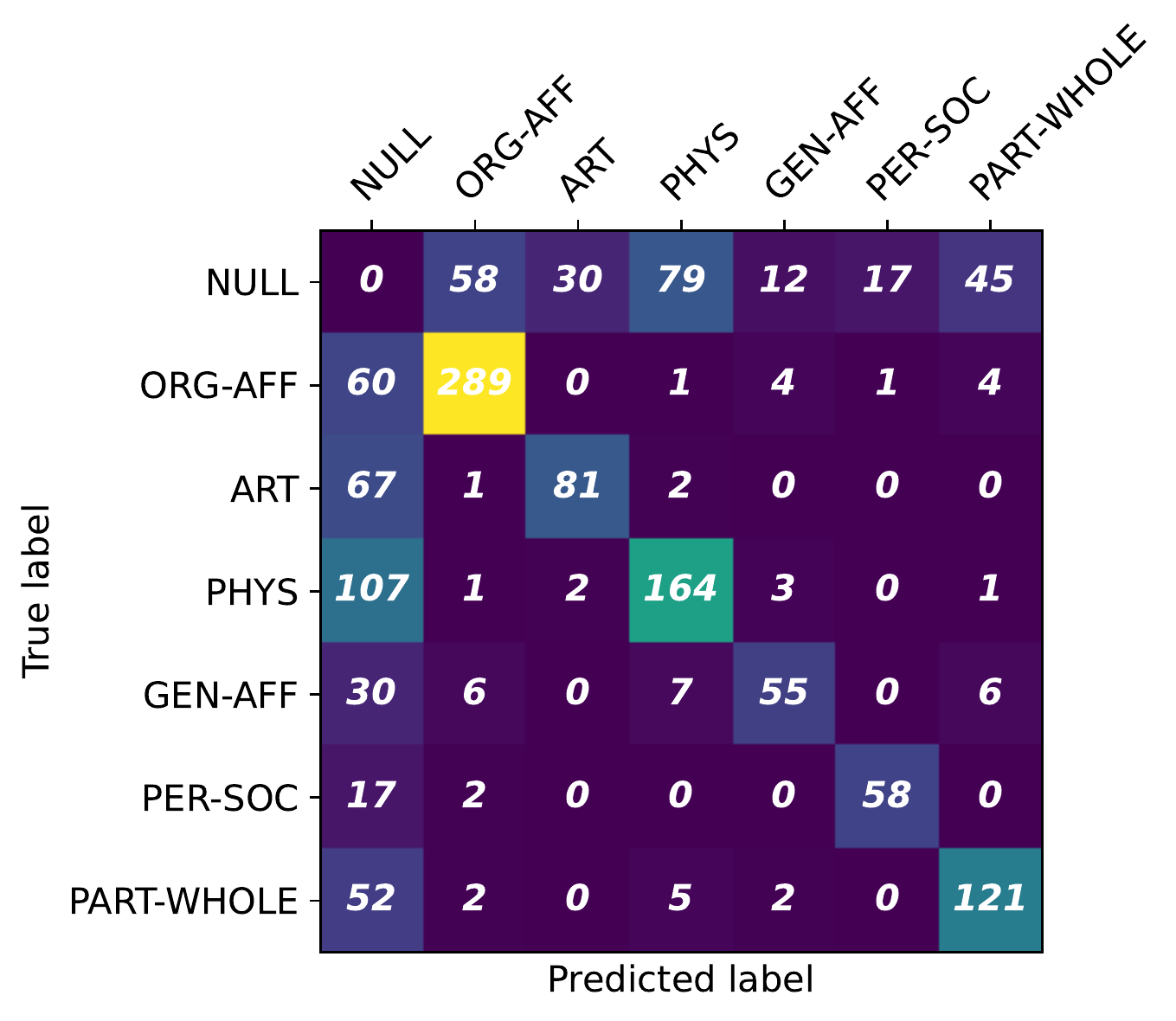}
        \end{minipage}
        }
    \subfigure[Error correction matrix]
        {                    
        \begin{minipage}[t]{0.9\columnwidth} 
        \centering                                       \includegraphics[width=\linewidth]{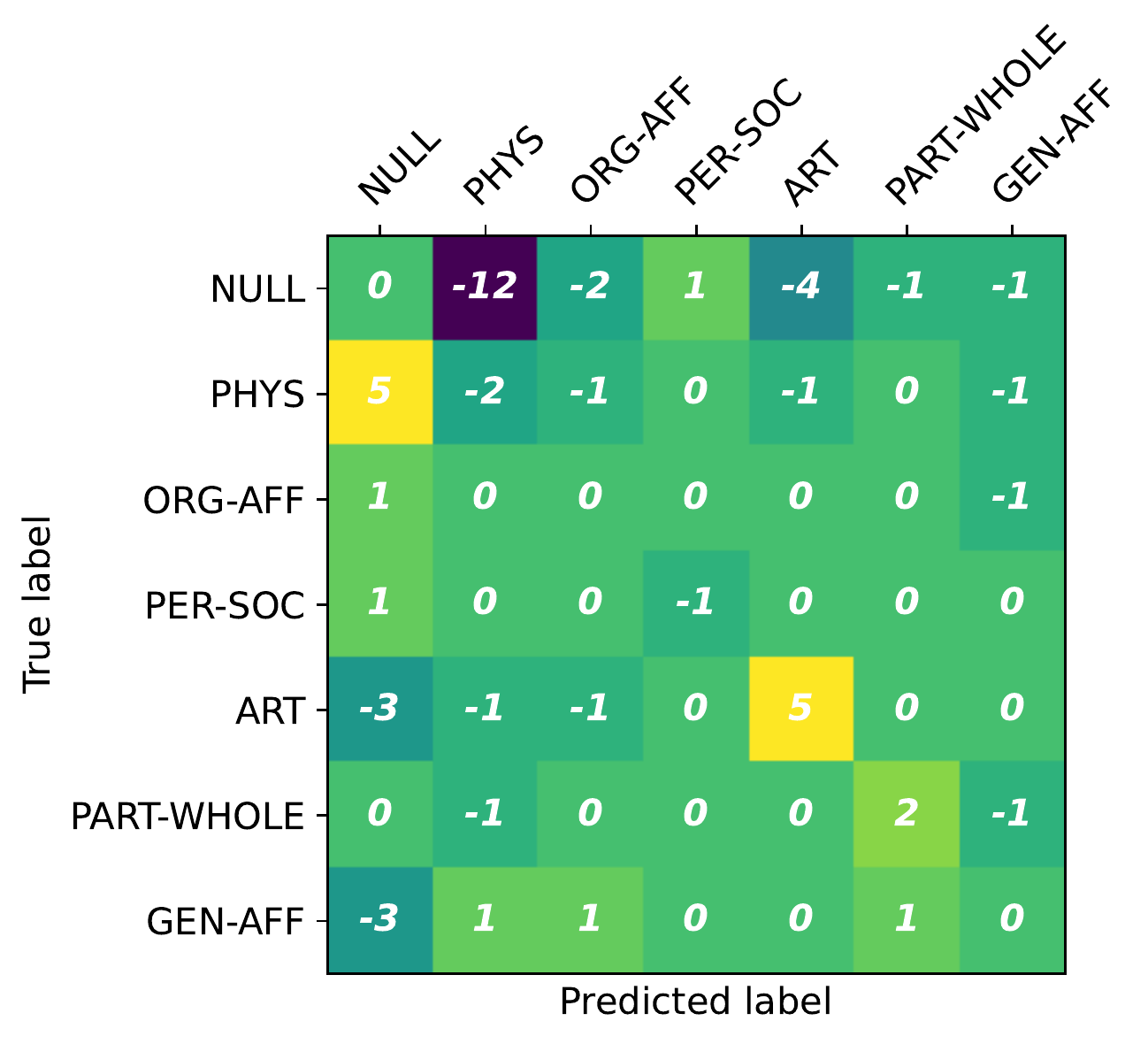}
        \end{minipage}
        }
\caption{Confusion matrix of the relation types. (a) The relation numbers of our baseline model on predicted entities. (b) The correction numbers of our high-order model relative to the baseline model. We do not have statistics on Null-Null.}
\label{fig:confusion}
\end{figure}

\section{Experimental results on ACE05-E given gold entities and triggers}
\label{sec:ace05-Egold}
Table \ref{apptab:goldace05e} shows the experimental results on ACE05-E given gold entities and triggers. We can find that without the error of the identification module, the performance gap between our baseline and high-order models further increases, and using both sibling factors and ternary factors improves further.


\section{Ablation Study}
\label{sec:ablation}
We show the experimental results of different factor combinations on Table \ref{apptab:ace05e}, Table \ref{apptab:ace05r} and Table \ref{tab:ace05eplusplus}.

\begin{table}[t!]
\centering
\setlength\tabcolsep{3pt}

\resizebox{.8\linewidth}{!}{%
\begin{tabular}{lcccc}
 & \textit{Ent} & \textit{Tri-C} & \textit{Arg-I} & \textit{Arg-C} \\ 
\hline
\textbf{CRFIE baseline}    & 96.0  & 93.1 & 70.7 & 68.3 \\
\textbf{CRFIE \textit{homo} (+sib)} & 96.0  & 93.6 & 72.0 & 69.2 \\
\textbf{CRFIE \textit{hete} (+ter)} & 95.9  & 94.1 & 71.7 & 69.2 \\
\textbf{CRFIE \textit{homo+hete} (+sib+ter)} & 96.0  & 93.6 & 72.3 & 69.4\\
\bottomrule
\end{tabular}
}
\caption{Average F1 on ACE05-E dataset. The gold triggers and entities are given.}
\label{apptab:goldace05e}
\vspace{-1em}
\end{table}

\begin{table}[t!]
\small
\centering
\setlength\tabcolsep{3pt}
\resizebox{\linewidth}{!}{
\begin{tabular}{lccccc}
 & \textit{Ent} & \textit{Tri-I} & \textit{Tri-C} & \textit{Arg-I} & \textit{Arg-C} \\ \hline

\hline
\textbf{CRFIE baseline}  & 90.8 & 77.7 & 74.8 & 58.5 & 56.4 \\
\textbf{CRFIE \textit{homo} (+sib)} & 90.6 & 77.7 & 74.5 & 59.1 & 57.1 \\
\textbf{CRFIE \textit{homo} (+sib+cop)} & 90.8 & 77.7 & 74.6 & 58.7 & 57.1 \\
\textbf{CRFIE \textit{hete} (+ter)} & 90.7 & 77.7 & 74.3 & 59.2 & 57.2 \\
\textbf{CRFIE \textit{homo+hete} (+sib+ter)} & 90.6 & 77.7 & 74.3 & 59.6 & 57.5 \\

\bottomrule
\end{tabular}
}
\caption{Average F1 on ACE05-E dataset with encoders of BERT-large-cased }
\label{apptab:ace05e}
\vspace{-0.5em}
\end{table}

\begin{table}[t!]
\centering
\setlength\tabcolsep{1.8pt}
\resizebox{.8\linewidth}{!}{%
\begin{tabular}{lccc}
 & \textit{Ent} & \textit{Rel} & \textit{Rel+} \\ \hline

\hline
\textbf{CRFIE baseline} & 89.8 & 69.9 & 67.5 \\ 
\textbf{CRFIE \textit{homo} (+sib)} & 90.0 & 70.8 & 68.1 \\
\textbf{CRFIE \textit{homo} (+cop)}  & 90.1 & 70.1 & 68.0 \\
\textbf{CRFIE \textit{homo} (+gp)}  & 90.2 & 70.0 & 67.7 \\
\textbf{CRFIE \textit{homo} (+sib+cop)}  & 90.2 & 70.8 & 68.2 \\
\textbf{CRFIE \textit{hete} (+ter)} & 90.1 & 70.4 & 68.3 \\
\bottomrule
\end{tabular}
}
\caption{Average F1 on ACE05-R dataset with encoder of ALBERT-XXLarg-v1}
\label{apptab:ace05r}
\vspace{-0.5em}
\end{table}



On Table \ref{tab:ace05eplusplus}, \textit{role-sib} represents sib of role pairs, \textit{rel-sib} represents sib of relation pairs, and \textit{r+r-sib} represents sib of both role pairs and relation pairs. The \textit{hete (+cop), hete (+gp), hete (+cop+gp)} are in \textbf{\textit{hete} case (i)}.

\begin{table}[t!]
\small
\centering
\setlength\tabcolsep{2pt}
\resizebox{\linewidth}{!}{
\begin{tabular}{lcccccc}
 & \textit{Ent} & \textit{Rel}& \textit{Tri-I} & \textit{Tri-C} & \textit{Arg-I} & \textit{Arg-C} \\ 
\hline 
\textbf{CRFIE baseline}                                           & 90.8         & 65.3         & 77.4           & 74.6           & 60.0           & 58.1            \\
\textbf{CRFIE \textit{homo} (role-sib)}                           & 90.8         & 65.1         & 77.4           & 74.6           & 60.3           & 58.4            \\
\textbf{CRFIE \textit{homo} (rel-sib)}                            & 91.0         & 65.6         & 77.4           & 74.8           & 60.1           & 58.5            \\
\textbf{CRFIE \textit{homo} (r+r-sib)}                            & 90.9         & 65.4         & 77.4           & 74.8           & 60.1           & 58.3            \\
\textbf{CRFIE \textit{hete} (+cop)}                             & 90.7         & 65.9         & 77.4           & 74.6           & 60.3           & 58.2            \\
\textbf{CRFIE \textit{hete} (+gp)}                              & 90.7         & 65.8         & 77.4           & 75.1           & 60.8           & 59.0            \\
\textbf{CRFIE \textit{hete} (+cop+gp)}                          & 90.7         & 65.1         & 77.4           & 74.8           & 60.3           & 58.5            \\
\begin{tabular}[c]{@{}l@{}}\textbf{CRFIE \textit{homo} case (i) }\\\textbf{~ ~ ~ + \textit{homo} case (ii)}\end{tabular} & 90.9         & 65.4         & 77.4           & 74.8           & 60.1           & 58.3            \\
\bottomrule
\end{tabular}
}
\caption{Average F1 on ACE05-E+ dataset. All models use BERT-large-cased encoder.}
\label{tab:ace05eplusplus}
\end{table}

\section{Error Correction Analysis}
\label{sec:error}
We take joint EntR and RelE as an example to show the number of error corrections of our high-order model compared to our baseline model in terms of relation types. From Fig. \ref{fig:confusion},
we can find that our high-order model corrects the errors of our baseline model in relation types (the numbers are expected to be positive in the diagonal and to be negative otherwise).

\section{Re-evaluation of PL-Marker}
\label{sec:reimp}
For the relation extraction task, some corpus have symmetric relations, meaning the ordering of the two entities does not matter~(e.g.,  `PER-SOC' in ACE2005). A symmetric relation is only annotated in one direction in the annotation data.
PL-Marker counts a symmetric relation twice both for prediction number and gold number, but other work only counts once for the prediction and gold numbers.

\end{document}